\documentclass[11pt]{article} % For LaTeX2e
\usepackage{rldmsubmit,palatino}
\usepackage{graphicx}
\usepackage{epigraph}

\title{Learned human-agent decision-making, communication\\ and joint action in a virtual reality environment}

\author{
Patrick M. Pilarski\thanks{Corresponding author. This work was conducted at DeepMind, with collaboration from the University of Alberta.}\\
DeepMind \& University of Alberta\\
Edmonton, Alberta, Canada\\
\And
Andrew Butcher \\
DeepMind \\
Edmonton, Alberta, Canada\\
\And
Michael Johanson \\
DeepMind \\
Edmonton, Alberta, Canada\\
\AND
Matthew M. Botvinick \\
DeepMind \\
London, UK\\
\And
Andrew Bolt\\
DeepMind \\
London, UK\\
\And
Adam S. R. Parker \\
University of Alberta\\
Edmonton, Alberta, Canada\\
}

\begin{document}

\maketitle

\begin{abstract}
Humans make decisions and act alongside other humans to pursue both short-term and long-term goals. As a result of ongoing progress in areas such as computing science and automation, humans now also interact with non-human agents of varying complexity as part of their day-to-day activities; substantial work is being done to integrate increasingly intelligent machine agents into human work and play. With increases in the cognitive, sensory, and motor capacity of these agents, intelligent machinery for human assistance can now reasonably be considered to engage in {\em joint action} with humans---i.e., two or more agents adapting their behaviour and their understanding of each other so as to progress in shared objectives or goals. The mechanisms, conditions, and opportunities for skillful joint action in human-machine partnerships is of great interest to multiple communities. Despite this, human-machine joint action is as yet under-explored, especially in cases where a human and an intelligent machine interact in a persistent way during the course of real-time, daily-life experience (as opposed to specialized, episodic, or time-limited settings such as game play, teaching, or task-focused personal computing applications). In this work, we contribute a virtual reality environment wherein a human and an agent can adapt their predictions, their actions, and their communication so as to pursue a simple foraging task. In a case study with a single participant, we provide an example of human-agent coordination and decision-making involving prediction learning on the part of the human and the machine agent, and control learning on the part of the machine agent wherein audio communication signals are used to cue its human partner in service of acquiring shared reward. These comparisons suggest the utility of studying human-machine coordination in a virtual reality environment, and identify further research that will expand our understanding of persistent human-machine joint action.
\end{abstract}


\acknowledgements{We are deeply indebted to our DeepMind colleagues Drew Purves, Simon Carter, Alex Cullum, Kevin McKee, Neil Rabinowitz, Michael Bowling, Richard Sutton, Joseph Modayil, Max Cant, Bojan Vujatovic, Shibl Mourad, Leslie Acker, and Alden Christianson for their support, suggestions, and insight during the pursuit of the work in this manuscript.}

\startmain

\epigraph{``First Thoughts are the everyday thoughts. Everyone has those. Second Thoughts are the thoughts you think about the way you think. People who enjoy thinking have those. Third Thoughts are thoughts that watch the world and think all by themselves.''}{\em Terry Pratchett, A Hat Full of Sky}

\vspace{-1.5em}
\section{Understanding and Improving Human-Machine Joint Action}

Humans regularly make decisions with and alongside other humans. In what has come to be defined as {\em  joint action}, humans coordinate with other humans to achieve shared goals, sculpting both their actions and their expectations about their partners during ongoing interaction (Sebanz et al. 2006; Knoblich et al. 2011; Pesquita et al. 2018). Humans now also regularly make decisions in partnership with computing machines in order to supplement their abilities to act, perceive, and decide (Pilarski et al. 2017). It is natural to expect that joint action with machine agents might be able to improve both work and play. In situations where someone is limited in their ability to perceive, remember, attend to, respond to, or process stimulus, a machine counterpart's specialized and complementary abilities to monitor, interpret, store, retrieve, synthesize, and relate information can potentially offset or even invert these limitations. Persistent computational processes that extend yet remain part of human cognition, perhaps best described in the words of the author Terry Pratchett as ``third thoughts,'' are evident in common tools like navigation software and calendar reminders.

Specifically with a focus on joint action, as opposed to more general forms of human-machine interaction, Moon et al. (2013), Bicho et al. (2010), Pilarski et al. (2013), Pezzulo et al. (2011), and others have provided compelling examples of fruitful human-machine coordination wherein a human and a machine work together and co-adapt in real-time joint action environments. One striking characteristic of many of these examples, and what separates them from other examples of human-machine interaction, is that they occur within {\em peripersonal space} (Knoblich et al. 2011)---i.e., interaction is perceived by the human to unfold continuously in the region of physical space surrounding their body and upon which they can act. While the perception of spatial and temporal proximity between partners has been shown to significantly influence joint decision making (as reviewed by Knoblich et al. 2011), peripersonal joint action settings have to date received less attention than other settings for human-machine interaction. The study of how different machine learning approaches impact human-machine joint action is even less developed, but in our opinion equally important. Our present work therefore aims to extend the discussion on how a human decision maker (here termed a {\em pilot}) and a machine learning assistant (termed a {\em copilot}) can learn together and make decisions together in service of a shared goal. We do so by defining a virtual reality environment to probe real-time joint action between humans and learning machines, and, using this environment, contribute a human-machine interaction case study wherein we demonstrate the kinds of changes that might be expected as a pilot interacts with different machine learning copilots in a shared environment. 

\begin{figure}[b]
    \centering
    \includegraphics[height=2in]{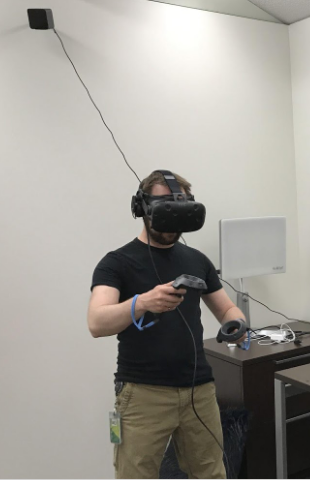}
    \includegraphics[height=2in]{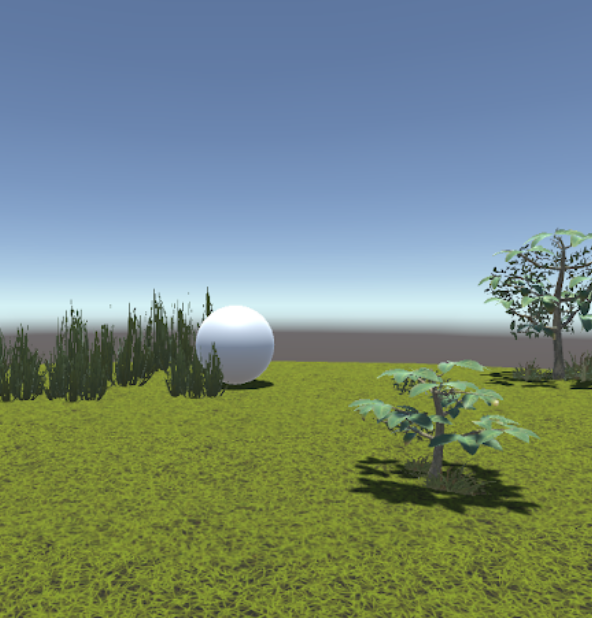}      
    \includegraphics[height=2in]{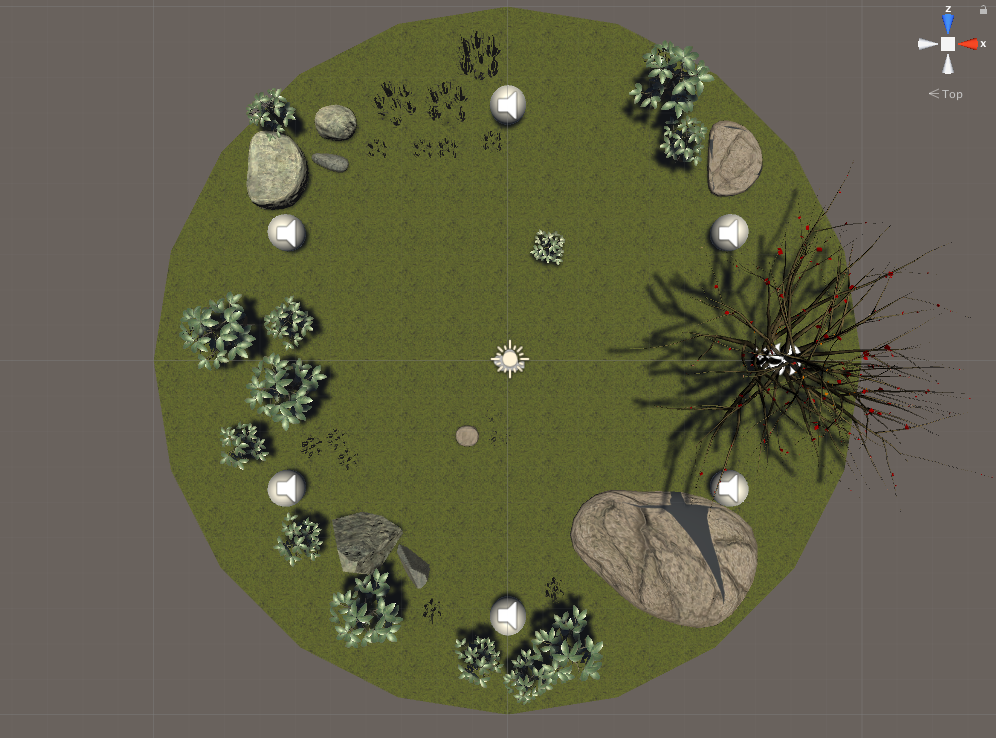}  
    \caption{\footnotesize The virtual-reality foraging environment used to explore human-agent learning and joint action, comprised of six equidistant fruit objects, background detail, and a repeated cycle of day and night illumination. The human participant (the {\em pilot}) was able to move about within the virtual world and used their hand-held controllers to both harvest fruit in varying states of ripeness and train their machine-learning assistant (the {\em copilot}).}
    \label{fig:world}
\end{figure}

\section{Virtual Reality Environment and Protocol}

A single participant engaged in a foraging task over multiple experimental blocks. This foraging task was designed so as to embed a hard-to-learn sensorimotor skill within a superficially simple protocol. In each block, the participant was asked to interact with a simulated world and a machine assistant via a virtual reality (VR) headset and two hand-held controllers (HTC Vive with deluxe audio strap). The pilot was instructed that in each block they were to move through the world to collect objects, and that these objects would grant them ``points''; they were told that, during the experiment, their total points would be reflected in visual changes to the environment, and that they would receive a unique, momentary audio cue whenever they gained or lost points as a result of their actions (and that they could also be given different audio cues in situations where they might expect to gain or lose points). 

\begin{figure}[t]
    \centering
    \includegraphics[width=5in]{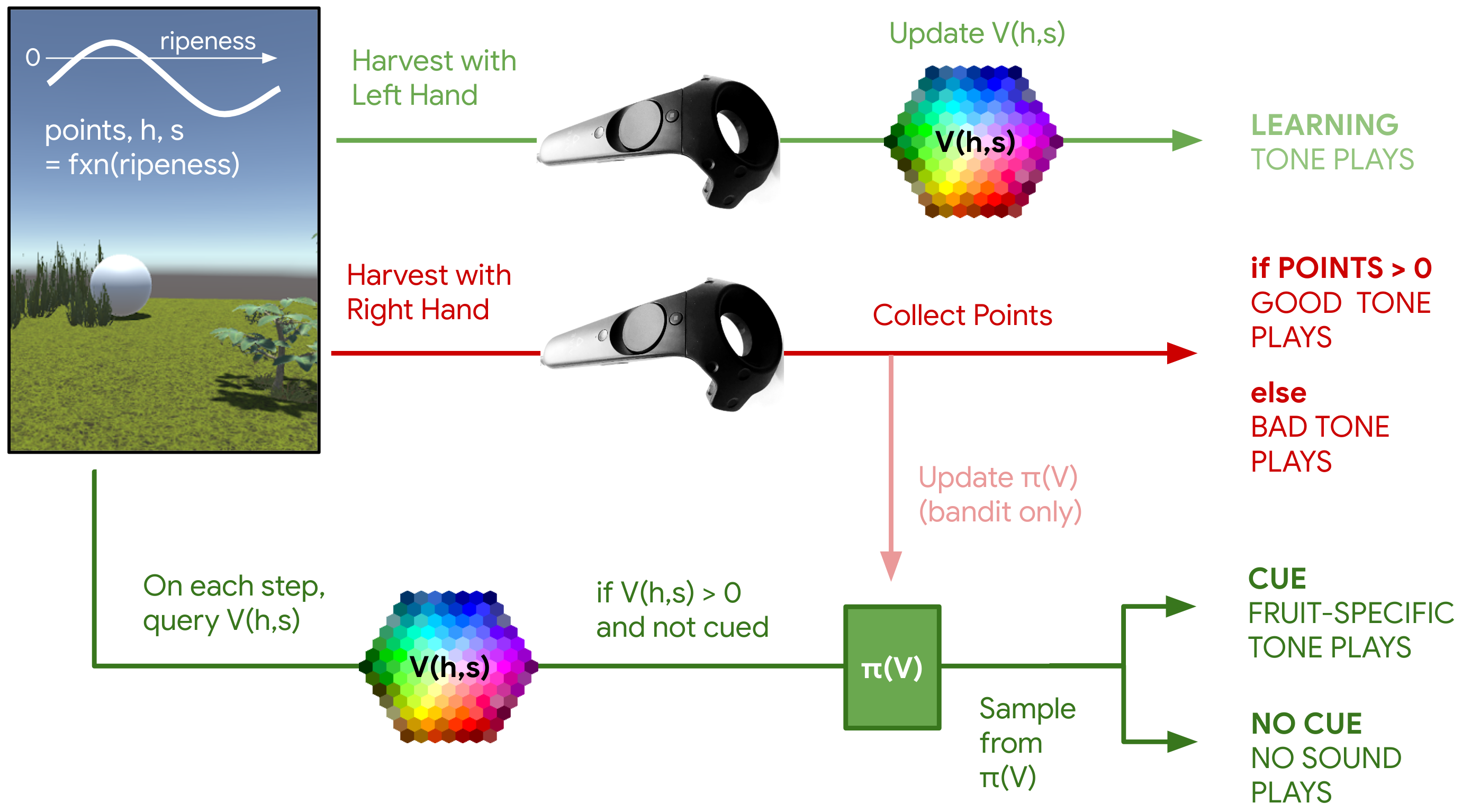}
    \caption{\footnotesize Schematic showing the interactions available to both the human pilot and the machine copilot. Using the right and left hand controllers, respectively, the pilot was able to harvest fruit or train their copilot to estimate the value of hue/saturation combinations, each action resulting in different sound cues as shown. On every time step, the copilot provided sound cues according to its learned predictions $V(h,s)$ and, for the bandit condition, its learned policy $\pi(V)$.}
    \label{fig:protocol}
    \vspace{-1em}
\end{figure}

Loosely inspired by the bee foraging example of Schultz et al. (1997), the virtual world presented to the pilot was a simple platform floating in space with six coloured balls (``fruit'') placed at equidistant locations around its perimeter (Fig. \ref{fig:world}). To tease out behavioural changes in different conditions for pilot/copilot joint action, over the course of a single trial the light level in the world varied from full sunlight (``day'') to regular periods of darkness lasting roughly 20s (``night'', c.f., Fig. \ref{fig:results}c, bottom trace). While copilot perception was unaffected by light level, during twilight and night phases the pilot was by design unable to determine the colour of any objects in the environment, seeing only their general shape. The main mechanic of the world was that the pilot could collect fruit to either increase their points or, in some blocks, collect fruit in order to teach their copilot. 

To create conditions for skill learning on the part of the pilot, the environment was designed with a confounding factor (ripeness) that related the appearance of the fruit as observed by the pilot to the reward structure of the task. Over the course of time, each fruit underwent a ``ripening cycle'' wherein it cycled through a progression of colours---hue and saturation levels---and reward that varied in relation to the time since the fruit's appearance. If a fruit was not contacted before a fixed time interval by the pilot, i.e., the end of the ripening cycle, it would disappear and no point gains or losses will be credited to the participant. A short, variable time after the participant collected a fruit or the fruit disappeared due to time, a new fruit would drop from above to the same position as the previous fruit. These new fruit were assigned a random fraction between 0\% and 95\% through their ripening cycle.  

The pilot started each trial in the centre of the platform, was able to move short distances and rotate in place, and could make contact with the fruit using either their left or right hand controllers (Fig. \ref{fig:protocol}). Upon making contact with one of the fruit via their right-hand controller (``harvesting''), the ball would disappear and the pilot would hear an audio cue indicating that they either gained or lost points (one distinct sound for each case, with a small mote of light appearing in place of the fruit when points were gained). Upon contacting a ball with their left-hand controller (``teaching''), the fruit would disappear with no points gained or lost by the pilot, and they would hear a unique tone indicating that they had given information to their copilot (for blocks involving a copilot, otherwise the fruit disappeared and no sound played).

Harvesting fruit was the only way the pilot could gain points. The points gained or lost by the participant for collecting a fruit was determined by a sinusoidal reward function that varied according to the time since the beginning of the ripening cycle (Fig. \ref{fig:protocol}). All fruit in a given trial were assigned the same, randomly selected reward sinusoid (in terms of phase) and related hue/saturation progression---i.e., all balls in a given trial would ripen in the same way, and would generate points in the same way, but these ripening mechanics would vary from trial to trial. The frequency of the reward sinusoid (i.e., number of reward maxima and minima until a fruit's disappearance), positive/negative reward offset, length of the ripening cycle, and time range until reappearance (all akin to difficulty), were predetermined and held constant across all trials and blocks. Difficulty in terms of these parameters was empirically preselected and calibrated so as to provide the pilot with a challenging pattern-learning problem that was still solvable within a single trial.

Following an extended period of familiarization with the navigation and control mechanics of the VR environment and the different interaction conditions, the participant (one of the co-authors for this pilot study) experienced three experimental blocks each consisting of three 180s trials; each trial utilized previously unseen ripening mechanics in terms of colour presentation and points phase. There was an approximately 20s break between trials.  The blocks related to the three different conditions (Fig. \ref{fig:protocol}), presented in order, as follows:

{\bf Condition 1, No copilot (NoCP): } The pilot harvested fruit without any signalling or support from a machine copilot. Using the left, training hand to contact fruit had no effect, and did not provide any additional audio cues.

{\bf Condition 2, Copilot communication via Pavlovian control (Pav):} The pilot harvested fruit with support from a machine copilot that {\em provided audio cues in a fixed way} based on its learned predictions. As described above, the pilot was able to train the copilot by contacting fruit with their left hand controller. Practically, this amounted to updating the value function of the copilot, denoted $V(h,s)$, according to the points value associated with a fruit's current hue $h$ and saturation level $s$ at the time of contact; updates were done via simple supervised learning.  At each point in time, the copilot queried its learned value function $V(h,s)$ with the current colour values of each of the six fruits, and, if the value of $V(h,s)$ was positive for a fruit, triggered an audio cue that was unique to that fruit---each fruit had a characteristic sound. In other words, feedback from the copilot to the human pilot was based on a pre-determined function that mapped learned predictions to specific actions (an example of Pavlovian control and communication, c.f., Modayil et al. (2014) and Parker et al. (2014)).

{\bf Condition 3, Copilot communication learned through trial and error (Bandit):}  The pilot harvested fruit with support from a machine copilot that {\em provided audio cues in an adaptable way} based on collected points (reward) and its learned predictions. This condition was similar to Condition 2 in terms of how the pilot was able to train the copilot. However, instead of deterministically playing an audio cue for the pilot each time the copilot's prediction $V(h,s)$ for a given fruit was positive, the copilot was instead presented with a decision whether or not to play an audio cue for the pilot. The decision to play a cue was based on a stochastic policy $\pi(V)$ that was updated as in a contextual bandit approach (Sutton and Barto, 2018) according to the points collected by the pilot if and when a fruit was harvested. In essence, if the copilot cued the user and this resulted in the pilot gaining points a short time later, it would reinforce the copilot's probability of playing a sound when it predicted a similar level of points in the future; should the pilot instead harvest the fruit and receive negative points, as when a fruit is harvested after the peak in its ripening cycle and/or the pilot is consistently slow to react to the copilot's cue, the copilot would decrease its probability of playing a sound when it predicted that level of expected points. The copilot's control policy used learned predictions as state, c.f., prediction in human and robot motor control (Wolpert et al. 2001; Pilarski et al. 2013).

\section{Results and Discussion}

\begin{figure}[t]
    \centering
     {\bf(a)} \includegraphics[height=1.4in]{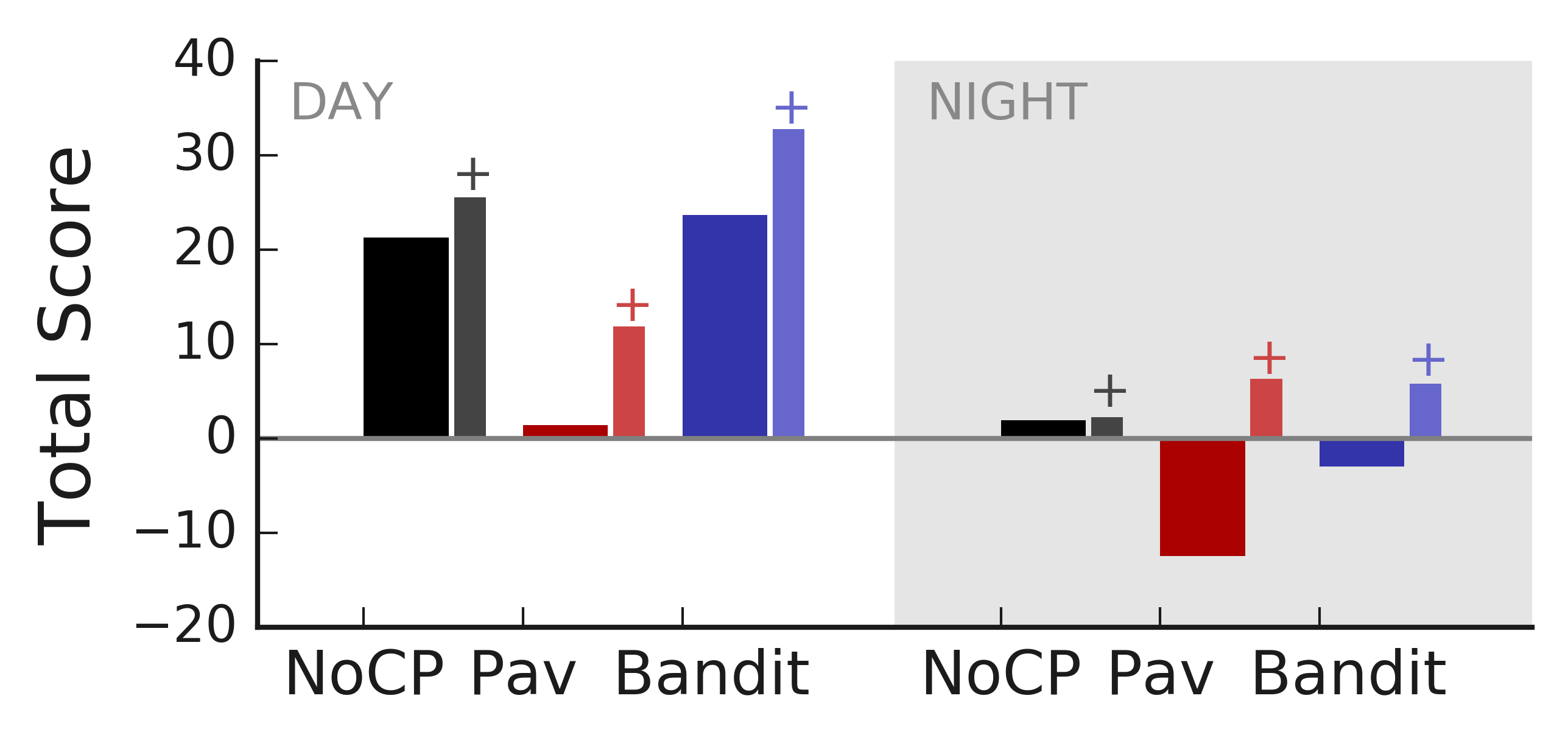} \hfil
     {\bf(b)} \includegraphics[height=1.4in]{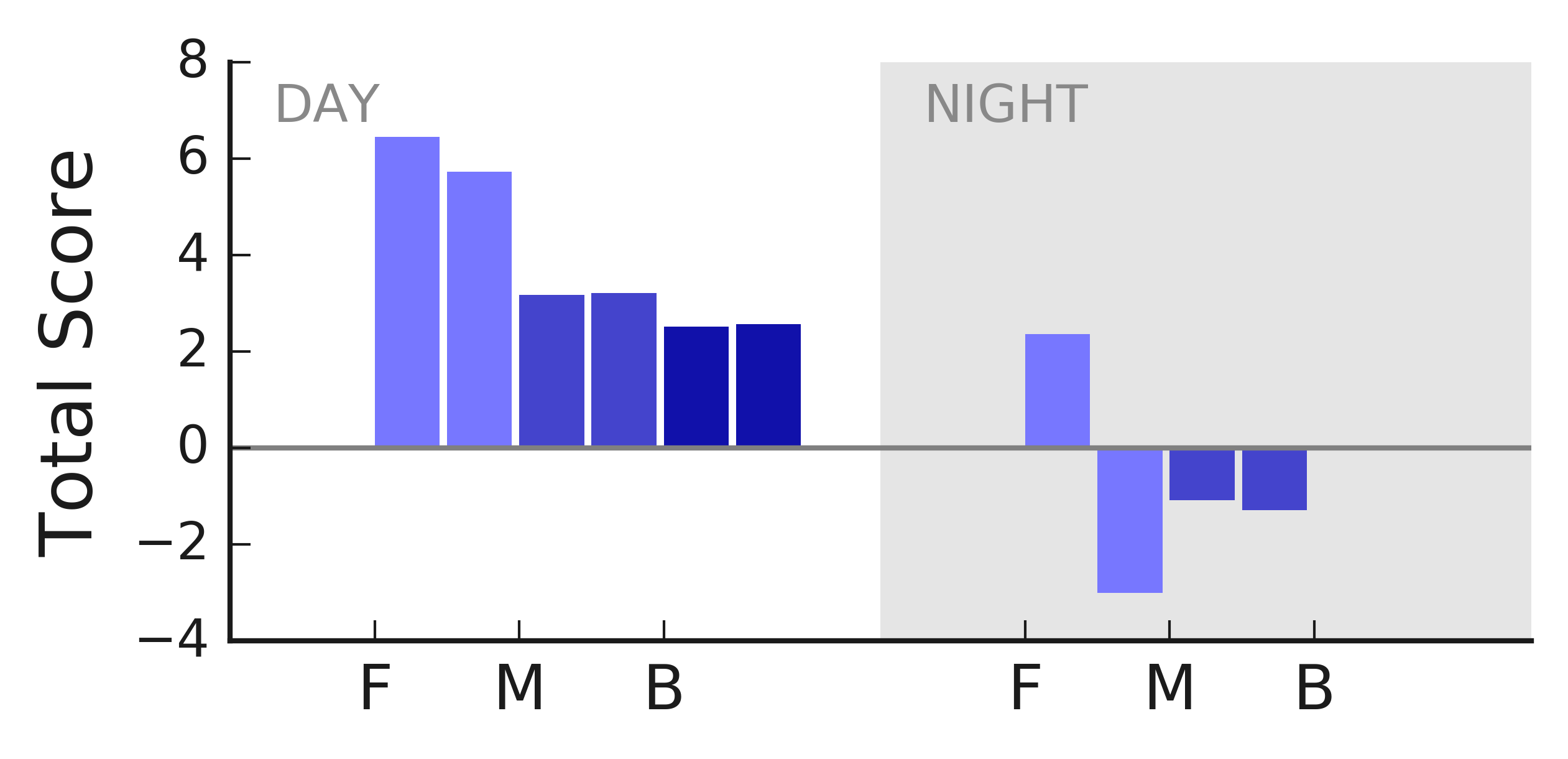}\\ 
 {\bf(c)} \includegraphics[width=6.6in]{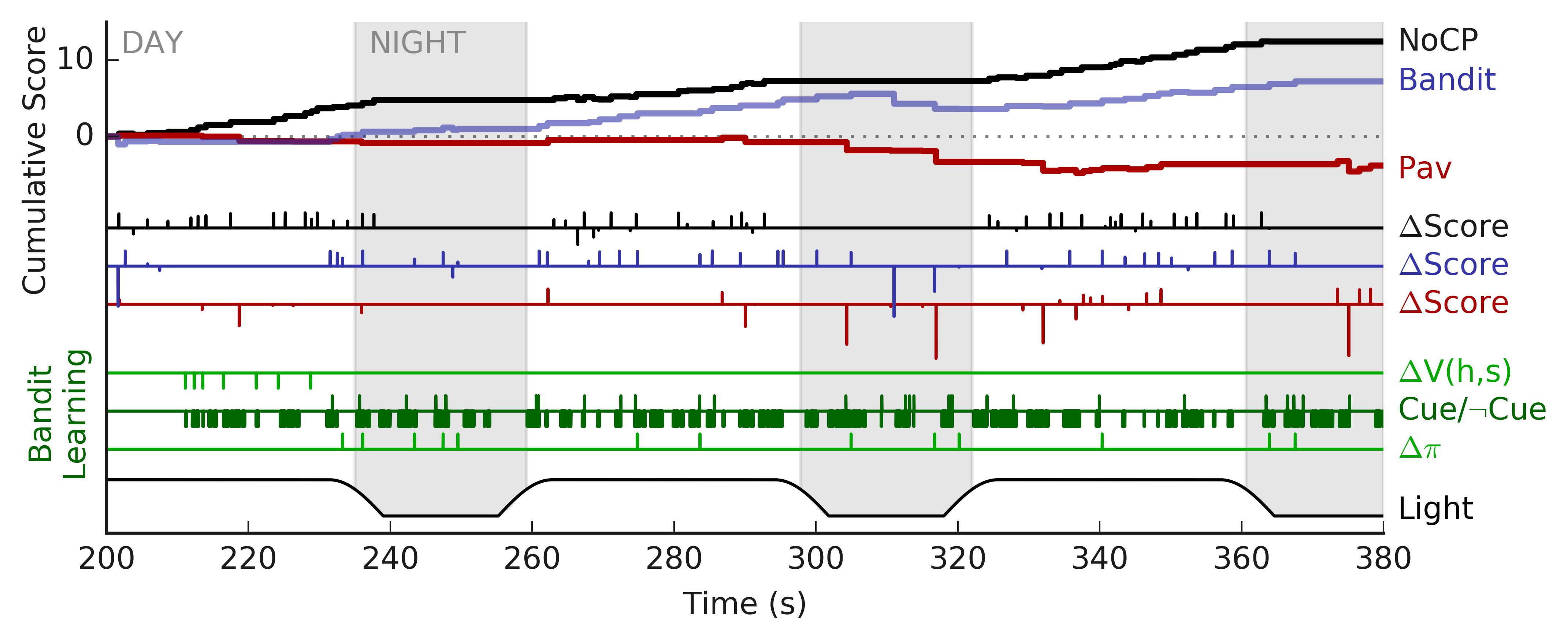}\\
    \caption{ \footnotesize Results from a single pilot participant for the three experimental conditions ({\em NoCP, Pav,} and {\em Bandit}). (a) Total score acquired by the pilot in each condition during day and night, as summed over all three trials in a block, along with the total score excluding any events with negative points (+); (b) total Bandit score with respect to fruit location to the front (F), middle (M) or back (B) with respect to the pilot's starting orientation; and (c) representative example of time-series data from the second trial for all three conditions, cross-plotting cumulative score, changes in score ($\Delta Score$), light level, and bandit learning in terms of human teaching actions ($\Delta V(h,s)$), copilot decisions to cue or not cue ({\em up/down ticks}), and post-cue updates to the copilot's policy as a result of pilot activity ($\Delta \pi$).}
    \label{fig:results}
    \vspace{-1em}
\end{figure}

Figure \ref{fig:results} presents the aggregate behaviour of the pilot in terms of total score over all three trials per condition, a fruit-by-fruit breakdown of total score, and a detailed presentation of time-series data from the second trial of the experiment. {As a key finding, we observed that interaction with different copilots (Pav and Bandit) led to different foraging behaviours on the part of the pilot during day and night, especially as compared to the no copilot (NoCP) condition.} Interactions with both the Pav and the Bandit copilot led to more foraging behaviour during night-time periods, as compared to the NoCP condition (Fig. \ref{fig:results}c). Learning to interpret the communication from these copilots appeared to induce multiple foraging mistakes on the part of the pilot, especially during night-time phases (Fig. \ref{fig:results}a,c) and less familiar fruit locations (Fig. \ref{fig:results}b). Despite this, the total points collected in the absence of any mistakes (the score without any negative point value events, Fig. \ref{fig:results}a+) suggest that collaboration with a policy-learning copilot could potentially lead to effective joint action once a good policy has been learned by both the pilot and the copilot. Teaching interactions ($\Delta V(h,s)$, Fig. \ref{fig:results}c) also provided a useful window into pilot skill learning. Broadly, the behaviour patterns observed in this preliminary study suggest that gradual addition of cues from a copilot, as in the Bandit policy-learning condition, is likely more appropriate than a strictly Pavlovian control approach. These initial results also indicate that there is room for more complex copilot architectures that can capture the appropriate timing of cues with respect to pilot activity (e.g., a pilot harvesting wrong fruit, or hesitation as per Moon et al. (2013)), and motivate more detailed study into the impact that pilot head position, gaze direction, light level, and other relevant signals have on a copilot's ability to generate good cues. Time delays and credit assignment matter in this joint-action setting and require further thought. 

{\bf Conclusions:} This work demonstrated a complete (though straightforward) cycle of human-agent co-training and learned communication in a VR environment, where closing the loop between human learning (human learns then trains an agent regarding patterns in the world) and agent learning (agent learns to make predictions and provide cues that must be learned by the pilot) appears possible to realize even during brief interactions. The VR fruit foraging protocol presented in this work proved to be an interesting environment to study pilot-copilot interactions in detail, and allowed us to probe the way human-agent behaviour changed as we introduced copilots with different algorithmic capabilities.

\section*{References}

\footnotesize

Bicho, E., et al. (2011). Neuro-cognitive mechanisms of decision making in joint action: A human-robot interaction study. {\em Human Movement Science} 30, 846--868.

Knoblich, G., et al. (2011). Psychological research on joint action: Theory and data. In WDK2003 (Ed.), {\em The Psychology of Learning and Motivation} (Vol. 54, pp. 59–-101). Burlington: Academic Press.

Modayil, J., Sutton, R. S. (2014). Prediction driven behavior: Learning predictions that drive fixed responses. {\em AAAI Wkshp. AI Rob}.

Moon, A., et al. (2013). Design and impact of hesitation gestures during human-robot resource conflicts. {\em Journal of Human-Robot Interaction} 2(3), 18--40.

Parker, A. S. R., et al. (2014). Using learned predictions as feedback to improve control and communication with an artificial limb: Preliminary findings. {\em arXiv}:1408.1913 [cs.AI]

Pesquita, A., et al. (2018). Predictive joint-action model: A hierarchical predictive approach to human cooperation. {\em Psychon. Bull. Rev.} 25, 1751--1769.

Pezzulo, G., Dindo, H. (2011). What should I do next? Using shared representations to solve interaction problems. {\em Exp. Brain. Res.} 211, 613--630.

Pilarski, P. M., et al. (2013). Real-time prediction learning for the simultaneous actuation of multiple prosthetic joints. {\em Proc. IEEE Int. Conf. Rehab. Robotics (ICORR)}, Seattle, USA, 1--8.

Pilarski, P. M., et al. (2017). Communicative capital for prosthetic agents. {\em arXiv}:1711.03676 [cs.AI]

Schultz, W., et al. (1997). A neural substrate of prediction and reward. {\em Science} 275(5306), 1593--9. 

Sebanz, N., et al. (2006). Joint action: Bodies and minds moving together. {\em Trends. Cogn. Sci.} 10(2), 70--76.

Sutton, R. S., Barto, A. G. (2018). {\em Reinforcement Learning: An Introduction}. Second Edition. Cambridge: MIT Press.

Wolpert, D. M., et al. (2001). Perspectives and problems in motor learning. {\em Trends Cogn. Sci.} 5(11), 487--494.

\end{document}